\documentclass{article}

\PassOptionsToPackage{numbers, sort&compress}{natbib}
\usepackage[main,preprint]{neurips_2026}
\usepackage[utf8]{inputenc}
\usepackage[T1]{fontenc}
\usepackage{xcolor}
\usepackage{hyperref}
\usepackage{url}

\usepackage{booktabs} 
\usepackage{amsfonts}
\usepackage{nicefrac}
\usepackage{microtype}
\usepackage{graphicx}
\usepackage{amsmath}
\usepackage{bm}
\usepackage{multirow}
\usepackage{wrapfig}
\hypersetup{
  colorlinks=true,
  linkcolor=cyan,
  urlcolor=cyan,
  citecolor=cyan
}

\title{LEIA: Learned Environment for Interactive Architected Materials}

\author{%
  Haiqian Yang, Yuan Cao, Markus J. Buehler \\
  Unreasonable Labs\\
  Mountain View, CA 94043, USA \\
  \texttt{\{haiqian, yuancao, mbuehler\}@unreasonablelabs.ai} \\
}

\begin{document}
\maketitle

\begin{abstract}
World models have enabled interactive exploration of game environments and robotic manipulation, but physical engineering remains beyond their reach: real materials exhibit nonlinear constitutive laws, carry history-dependent internal state, undergo inertial dynamics, and may possess hierarchical structures spanning multiple length scales. We present \textbf{LEIA} (\textbf{L}earned \textbf{E}nvironment for \textbf{I}nteractive \textbf{A}rchitected materials), a world model that lets engineers apply boundary conditions step by step and observe the resulting deformation and stress fields in real time. LEIA handles large three-dimensional unstructured meshes and generates autoregressive responses to user-specified loading. We introduce MicroPlate, a benchmark of architected plates spanning two regimes of microstructure modeling: architected lattices that resolve microstructure explicitly through three-dimensional geometry, and a homogeneous plate where microstructural change is modeled implicitly through internal degrees of freedom. MicroPlate is used to assess LEIA alongside four baseline methods across both regimes. Finally, we demonstrate that LEIA enables efficient candidate generation and ranking for fast surrogate-guided search for de novo designs of architected materials, with stress-accurate candidate ranking validated by finite element ground truth.
\end{abstract}

%-------------------------------------------------------------------
%                               INTRODUCTION
%-------------------------------------------------------------------
\section{Introduction}

\begin{wrapfigure}{r}{0.6\textwidth}
  \centering
  \vspace{-10pt}
  \includegraphics[width=0.6\textwidth]{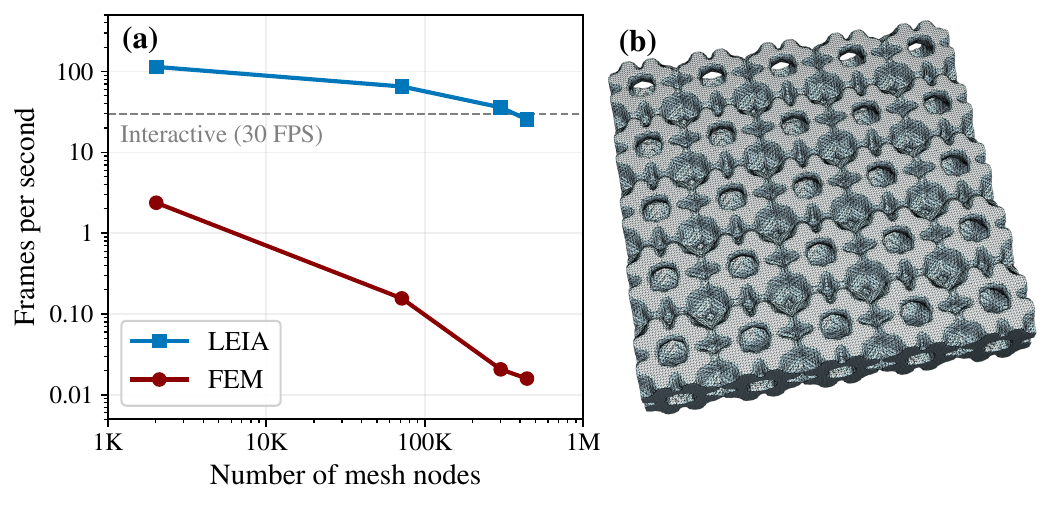}
  \caption{\footnotesize (a) Inference throughput of LEIA and FEM vs.\ mesh size. LEIA achieves high per-step inference speedup. Dashed line: interactive threshold (30 FPS). (b) A MicroPlate lattice plate (301,565 nodes).}
  \label{fig:teaser}
  \vspace{-10pt}
\end{wrapfigure}

Hierarchical materials, structures whose property emerges from geometry organized across multiple length scales, are among the most effective designs in nature \cite{meyers2008biological, wegst2015bioinspired}. Bone, nacre, and wood achieve combinations of strength, toughness, and light weight that far exceed what their base constituents alone can provide. Inspired by these systems, engineered counterparts, commonly termed architected materials, now replicate and extend these principles across a wide range of applications \cite{portela2025enabling, xia2022responsive}: lattice-based implants match bone stiffness \cite{wang2016topological}, crash structures absorb energy through controlled strut buckling \cite{tancogne2016additively}, and metamaterials achieve programmable stiffness and tunable energy absorption through unit-cell design \cite{jiao2023metamaterials, shaikeea2022toughness}. Additive manufacturing has made these geometries fabricable, shifting the bottleneck from fabrication to design \cite{zheng2023deeplearning, bastek2025physicsinformed, zheng2026diffumeta}: given the vast space of topologies, strut dimensions, and base materials, how does one efficiently explore the geometry-property landscape to find structures that meet target performance?

The conventional answer is finite element method (FEM), but for architected materials the cost is prohibitive. Resolving detailed features in a single 3D lattice structure requires hundreds of thousands of tetrahedral elements \cite{hirschler2024reduced, guillet2026lattice}, and each nonlinear numerical solve is computationally expensive, even for moderate meshes \cite{gongora2024accelerating,jain2024latticegraphnet}. Meanwhile, navigating the combinatorial design space demands massive simulation campaigns: recent data-driven studies have required extensive finite element evaluations to map the space of lattice topologies and geometric parameters \cite{bastek2022inverting, bastek2023inverse, zheng2023unifying}. Neural surrogates, neural network models that approximate complex physics simulations, can accelerate individual evaluations \cite{white2019multiscale, deng2022inverse}, and GPU-accelerated solvers reduce per-solve wall time \cite{xue2023jaxfem}, but interactive exploration of the design space remains an open problem.

World models, learned simulators that let agents interact with an environment through actions and observations, have transformed robotics by replacing expensive ground-truth simulators with fast neural surrogates \cite{ha2018worldmodels, hafner2023dreamerv3, yang2024unisim}. We propose to bring the same paradigm to materials engineering: an engineer applies boundary conditions interactively and observes the resulting deformation and stress in real time, accelerating the design loop by orders of magnitude. Building such an environment for architected materials requires a neural surrogate that handles large three-dimensional unstructured meshes, predicts both displacement and stress accurately under complex constitutive laws, and responds autoregressively to user-specified boundary conditions.

Neural surrogates have made substantial progress on mesh-based prediction. The community has converged on an encode-process-decode architecture rooted in Perceiver IO \cite{jaegle2022perceiverio,buehler2022fieldperceiver}: cross-attention compresses an arbitrary mesh to fixed-size latent tokens, a transformer processes them, and cross-attention decodes back to physical space \cite{wu2023lsm, alkin2024upt, serrano2024aroma, wang2024lno}. These models achieve high accuracy on established benchmarks and run much faster than solvers. Yet nearly all progress has been concentrated in fluid dynamics and largely confined to two-dimensional settings. Three-dimensional nonlinear solid mechanics, the domain most relevant to architected materials design, remains largely unexplored, with existing benchmarks limited to small meshes, static loading, or linear elasticity \cite{li2023geofno, pfaff2021mgn, jain2024latticegraphnet}.

Beyond mesh handling, an interactive environment for architected materials design must provide accurate stress fields. Engineers rely on stress for failure prediction and force-displacement characterization, and computing stress from displacement requires evaluating a constitutive law that maps deformation history to internal forces. For simple hyperelastic materials, stress can be recovered post-hoc by numerical differentiation of the predicted displacement field \cite{dagli2025vomp}. For materials with explicit microstructure resolved in the mesh, this approach becomes expensive at large mesh sizes. For materials with internal state variables that encode deformation history, displacement snapshots alone do not determine stress: identical current displacements can produce different stress fields depending on the loading path. We introduce a \emph{stress head}, a lightweight output branch that predicts stress directly, and evaluate it across both regimes on MicroPlate (Section~\ref{sec:results}).

An interactive environment must also respond to unseen boundary conditions at each timestep and quickly render the new states of the system. In game-world models, action conditioning lets agents interact with the environment step by step \cite{bruce2024genie, ha2018worldmodels, hafner2023dreamerv3}. We adopt the same mechanism for loading actions: at each timestep the user specifies a loading action, and LEIA conditions its dynamics on that action \cite{peebles2023dit,perez2018film}.

\begin{figure}[h!]
  \centering
  \includegraphics[width=\textwidth]{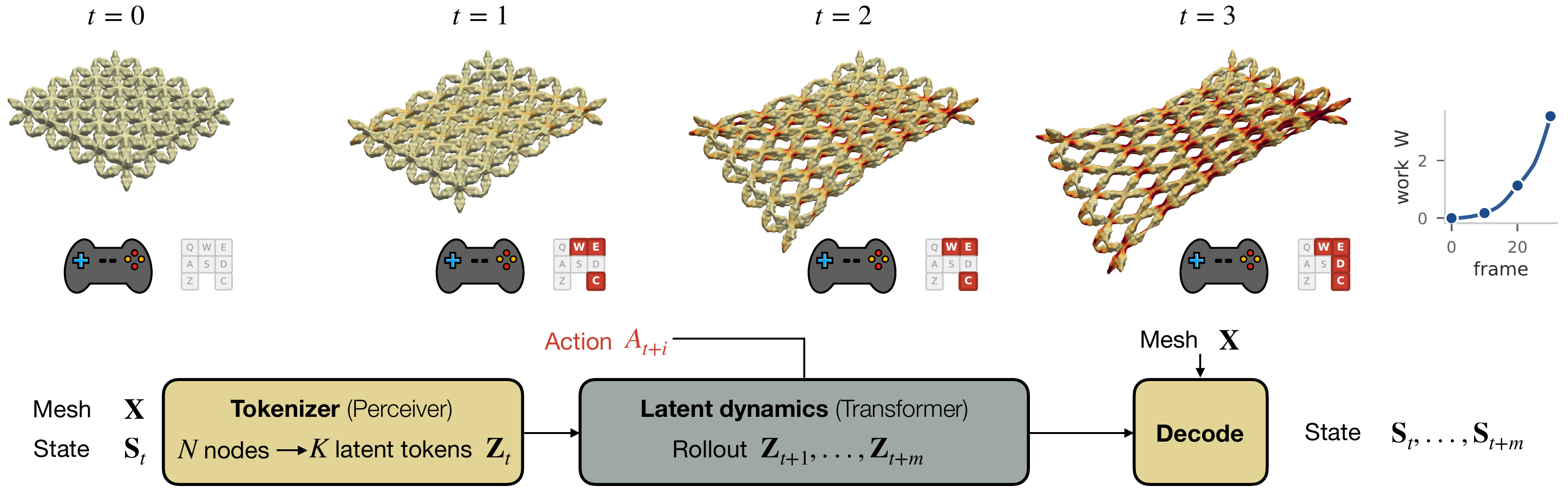}
  \caption{\footnotesize \textbf{LEIA architecture.} The tokenizer compresses the physical fields $\mathbf{S}_t$ on an arbitrary mesh $\mathbf{X}$ into $K$ latent tokens $\mathbf{Z}_t$ via Perceiver cross-attention. The dynamics transformer, conditioned on the boundary condition action $\mathbf{A}_t$, predicts the next latent state autoregressively. The decoder reconstructs the physical states at mesh nodes.}
  \label{fig:architecture}
\end{figure}

We present LEIA (\textbf{L}earned \textbf{E}nvironment for \textbf{I}nteractive \textbf{A}rchitected materials)~\footnote{Code is avaiable at \url{https://github.com/HaiqianYang-MechE/leia}}, an interactive, action-conditioned world model for architected materials. LEIA combines Perceiver tokenization for mesh-size invariance, action-conditioned transformer dynamics for boundary condition responsiveness, and the stress head for accurate stress readout. We evaluate LEIA on MicroPlate, a new benchmark covering 63 architected lattices with meshes of 71,000 to 442,000 nodes (two orders of magnitude larger than existing solid mechanics benchmarks) together with a visco-hyperelastic plate that exposes path-dependent stress. Our contributions:

\begin{itemize}
  \item LEIA, an interactive, action-conditioned world model for architected materials, enabling real-time design exploration under nonlinear mechanics with user-controlled loading.
  \vspace{-5pt}
  \item The stress head, a lightweight output branch applicable across architectures that provides accurate stress readout at constant cost regardless of constitutive complexity.
  \vspace{-5pt}
  \item MicroPlate, a two-regime benchmark of architected plates: 3D architected lattices where microstructure is modeled explicitly through geometry, and a visco-hyperelastic plate where microstructural change is modeled implicitly through internal degrees of freedom.
  \vspace{-5pt}
  \item Surrogate-guided search for de novo architected materials design, where beam search with LEIA rapidly evaluates candidates and discovers improved designs, validated by FEM.
\end{itemize}

\section{Related Work}

\paragraph{Neural operator architectures.}
Neural surrogates for PDEs have converged on a shared encode-process-decode template rooted in Perceiver IO \cite{jaegle2022perceiverio}: cross-attention compresses an arbitrary mesh to fixed-size latent tokens, a transformer processes them, and cross-attention decodes back to the physical domain. LSM \cite{wu2023lsm}, UPT \cite{alkin2024upt}, AROMA \cite{serrano2024aroma}, and LNO \cite{wang2024lno} each instantiate this pattern with different preprocessing and latent-dynamics choices. Physics-attention architectures have scaled to million-node geometries \cite{luo2025transolver++}. LEIA adopts the same template with Perceiver tokenization, FiLM-conditioned dynamics for boundary condition control, and a stress head for direct stress readout.

\paragraph{Autoregressive rollout and boundary condition conditioning.}
Most existing solid-mechanics benchmarks are quasi-static, but real materials exhibit rate-dependent behavior requiring long-horizon autoregressive rollout where error accumulates. Recent works address this via unrolled training \cite{list2024unrolled}, Hamiltonian inductive biases \cite{hoang2026igns}, and multi-wave architectures \cite{zhang2024sinenet}.
To respond to user-specified boundary conditions, standard surrogates such as MeshGraphNets \cite{pfaff2021mgn} rely on boundary node clamping, but clamping cannot propagate new loading conditions across the mesh in a single step, particularly when the system starts from rest. Explicit conditioning on the boundary action provides this information directly at every layer.

\paragraph{Derivative accuracy in neural surrogates.}
Stress is derived from the displacement gradient, so a surrogate that reconstructs displacement accurately does not necessarily produce accurate stress. Sobolev training \cite{czarnecki2017sobolev, cho2025sobolev, oh2026sobolevaccel,olearyroseberry2025difno} addresses the gap by supervising derivatives alongside function values, and \citet{bhattacharya2025memory} show that history-dependent constitutive laws require explicit encoding of the deformation path. In practice, DE-DeepONet \cite{qiu2024dedeeponet} and gPINNs \cite{yu2022gpinns} add derivative losses via automatic differentiation, but autograd incurs substantial overhead for high-dimensional outputs such as the 3D displacement Jacobian, compounding further with pushforward training. An alternative, established in molecular dynamics \cite{gasteiger2021gemnet} and neural radiance fields \cite{verbin2022refnerf}, is to predict derivatives directly via a lightweight output head. LEIA's stress head follows this pattern.

\paragraph{Stress prediction in solid mechanics.}
Engineering design requires accurate stress, not just displacement: stiffness, failure, and energy absorption all depend on quantities derived from the constitutive law and the displacement gradient.
A growing body of work predicts stress fields with neural networks: CNN approaches \cite{nie2020stress, mianroodi2021teaching}, FNO-based methods for composites \cite{rashid2022fno_stress} and viscoplastic materials \cite{khorrami2023viscoplastic, abueidda2021deep}, DeepONet branches for elastoplastic stress \cite{he2023deeponet_stress, he2024geom_deeponet}, LatticeGraphNet for Neo-Hookean lattice structures \cite{jain2024latticegraphnet}, GINOT for von Mises stress via transformer decoding \cite{liu2025ginot}, and comprehensive 3D benchmarks \cite{zhong2025fc4no}. PeFNO \cite{khorrami2024pefno} encodes the divergence-free constraint via a stress potential but is restricted to 2D regular grids. The MeshGraphNet-Transformer \cite{iparraguirre2026mgn_transformer} handles large-scale plasticity and impact by predicting full internal variable state. At the constitutive modeling level, Constitutive Artificial Neural Networks \cite{linka2023cann, tacke2026llm_cann} encode thermodynamic constraints to learn the material law. Concurrent work evaluates stress in crash simulation via geometric pre-training \cite{wu2026geopt} and graph/attention surrogates \cite{nabian2025crash}. Two gaps remain. First, none of these methods combine large 3D unstructured meshes with temporal, action-conditioned rollout. State-of-the-art 3D surrogates \cite{holzschuh2025p3d} achieve impressive scale on fluids, and adaptive mesh quantization \cite{dool2025amq} has begun incorporating hyperelasticity, yet 3D solid mechanics on unstructured meshes remains unaddressed. Second, none systematically evaluate how stress accuracy degrades as constitutive complexity increases.

%----------------------------------------------------------
%                           METHODS
%----------------------------------------------------------
\section{Methods}

\subsection{Problem formulation}
We consider finite deformation solid mechanics on a body occupying reference configuration $\Omega_0$. The displacement field $\mathbf{u}(\mathbf{X}, t)$ maps material points $\mathbf{X}$ to deformed positions $\mathbf{x} = \mathbf{X} + \mathbf{u}$, with deformation gradient $\mathbf{F} = \mathbf{I} + \nabla\mathbf{u}$ and right Cauchy-Green tensor $\mathbf{C} = \mathbf{F}^\top\mathbf{F}$. The material response is governed by a free energy $\Psi_\mathrm{R} = \Psi_\mathrm{R}^\mathrm{eq}(\bar{\mathbf{C}}) + \sum_{i} \Psi_\mathrm{R}^\mathrm{neq}(\bar{\mathbf{C}}, \mathbf{A}^{(i)}) + \Psi_\mathrm{R}^\mathrm{vol}(J)$, where $\bar{\mathbf{C}} = J^{-2/3}\mathbf{C}$ is the distortional Cauchy-Green tensor, $J = \det\mathbf{F}$, and $\mathbf{A}^{(i)}$ are symmetric positive-definite tensorial internal variables encoding deformation history across $N_\mathrm{b}$ visco-hyperelastic branches. The first Piola stress is $\mathbf{P} = 2\mathbf{F}\,\partial\Psi_\mathrm{R}/\partial\mathbf{C}$ and the Cauchy stress is $\bm{\sigma} = J^{-1}\mathbf{P}\mathbf{F}^\top$. The displacement field satisfies the balance of linear momentum,
\begin{equation}
  \mathrm{Div}\,\mathbf{P} = \rho_0\,\ddot{\mathbf{u}},
  \label{eq:momentum}
\end{equation}
with reference density $\rho_0$. Each internal variable evolves toward thermodynamic equilibrium via $\dot{\mathbf{A}}^{(i)} = \tau^{(i)^{-1}}(\bar{\mathbf{C}} - \mathbf{A}^{(i)})$ with $\det\mathbf{A}^{(i)} = 1$. Together these define a coupled initial-boundary-value problem: the displacement $\mathbf{u}$ and internal variables $\{\mathbf{A}^{(i)}\}$ co-evolve, and stress at any point depends on the full deformation history through $\mathbf{A}^{(i)}$. 
In the elastic limit ($N_\mathrm{b} = 0$, $\rho_0\ddot{\mathbf{u}} \to \mathbf{0}$), the system reduces to quasi-static hyperelasticity, where stress depends only on the current deformation gradient.

\paragraph{Datasets.}
MicroPlate (a benchmark for architected plates with explicit and implicit microstructure modeling) covers two regimes of modeling microstructural changes, both loaded via boundary condition actions (a mixture of 4 standard plate loading axes, i.e. stretch, twist, shear in two directions) with values drawn from $\{-1, 0, +1\}$ at each timestep for tractable FEM data generation. The lattice regime resolves microstructure explicitly through 3D geometry: 63 architected lattices with meshes of 71,000 to 442,000 nodes (8,836 trajectories of 30 quasi-static steps, Neo-Hookean material), where complex stress concentrations at strut junctions distinguish these plates from the homogeneous plate. The viscoelastic regime models microstructural change implicitly through internal degrees of freedom: a homogeneous plate (363 nodes, 1,000 trajectories of 100 action steps) under finite viscoelasticity with inertia (Arruda-Boyce equilibrium + 3 non-equilibrium branches). Each visco-hyperelastic trajectory applies 50 random loading steps then the exact reverse 50 unloading steps.
The two regimes vary along complementary axes: geometry varies in the lattice regime with the constitutive law held fixed; constitutive complexity varies in the viscoelastic regime with geometry held fixed. Full constitutive laws, material parameters, and FEM solver details are in Appendix~\ref{app:constitutive}.

\subsection{Model architecture.}
LEIA combines three design choices that address the requirements identified above (Figure~\ref{fig:architecture}). First, Perceiver cross-attention \cite{jaegle2022perceiverio} compresses fields on meshes of arbitrary size into a fixed-length latent representation, decoupling downstream model cost from mesh resolution. Second, action-conditioning \cite{perez2018film} injects the user's boundary condition action at every transformer layer, ensuring the dynamics model responds to new loading at each step. Third, a stress head, a single linear projection from the shared latent to six Cauchy stress components, provides stress readout at constant cost regardless of constitutive law complexity. 
Because stress and displacement share the same latent representation, we hypothesize that stress supervision forces the encoder to capture constitutive state that displacement alone cannot encode; Section~\ref{sec:plate_benchmarks} presents evidence consistent with this.

LEIA is trained in two stages: a \emph{tokenizer} is first trained, that compresses displacement and stress fields defined on arbitrary meshes into a fixed-size latent representation, followed by a second stage which trains a \emph{latent dynamics model} that predicts the next latent state conditioned on the current state and boundary condition action.

\paragraph{Tokenizer.}
The tokenizer follows the Perceiver IO encode-decode template \cite{jaegle2022perceiverio}, with encoder $\mathcal{E}_\phi$ and decoder $\mathcal{D}_\psi$:
\begin{equation}
  \mathbf{z} = \mathcal{E}_\phi(\mathbf{u}, \mathbf{X}), \qquad (\hat{\mathbf{u}}_j, \hat{\bm{\sigma}}_{\mathrm{sym},j}) = \mathcal{D}_\psi(\mathbf{z}, \mathbf{X}_j).
  \label{eq:tokenizer}
\end{equation}
A learned set of $K$ latent queries cross-attends into the $N$ mesh node features to produce a fixed-size latent representation $\mathbf{z} \in \mathbb{R}^{K \times H}$, refined by $L_\mathrm{enc}$ self-attention layers. Queries at arbitrary spatial positions $\{\mathbf{X}_j\}$ decode $\mathbf{z}$ back through cross-attention into the latents and two parallel linear heads producing displacement $\hat{\mathbf{u}}_j \in \mathbb{R}^3$ and Cauchy stress $\hat{\bm{\sigma}}_{\mathrm{sym},j} \in \mathbb{R}^6$. Node features project per-node displacement to dimension $H$ and add a Fourier positional encoding of the reference coordinate. The input displacement can be a single frame $\mathbf{u}^{(t)} \in \mathbb{R}^3$ or a temporal window of $F$ consecutive frames in $\mathbb{R}^{3F}$; multi-frame input provides velocity and acceleration context, beneficial for viscoelastic systems where internal state depends on deformation history. When the full mesh fits in GPU memory during training (e.g., the visco-hyperelastic plate), $L_\mathrm{dec}$ decoder self-attention layers attend across query positions to improve reconstruction accuracy; for large meshes requiring subsampled training (the lattice regime), we remove these layers to preserve resolution invariance (Appendix~\ref{app:hyperparams}).

\paragraph{Latent dynamics model.}
A transformer \cite{vaswani2017attention} predicts the next latent state from the current state and the boundary condition action:
\begin{equation}
  \mathbf{z}_{t+1} = f_\theta(\mathbf{z}_t,\;\mathbf{a}_t).
  \label{eq:dynamics}
\end{equation}
$f_\theta$ is a stack of $L_\mathrm{dyn}$ DiT-style adaLN blocks \cite{peebles2023dit, perez2018film}: each block applies action-conditioned scale and shift to the LayerNorm output before self-attention and again before the pointwise feedforward, with residual connections in the standard pre-norm transformer pattern. The four modulation parameters per block come from a per-block linear projection of a shared action embedding. The action $\mathbf{a}_t \in \mathbb{R}^4$ is the continuous-valued cumulative boundary condition at frame $t$; it enters at every block, ensuring the dynamics respond to boundary-condition changes immediately.

At inference the model operates autoregressively: starting from $\mathbf{z}_0$ (the encoding of the initial displacement), each predicted latent state is fed back as input, with the user supplying a new action $\mathbf{a}_t$ at each step. The decoder reconstructs displacement and stress from any intermediate $\mathbf{z}_t$.

\paragraph{Training procedure and implementation details.}
Details of the training procedure are shown in~\ref{app:hyperparams}. The Perceiver tokenizer is trained with a combined loss to reconstruct both the displacement field and the stress tensor field; the Transformer dynamics model to predict the next latent state conditioned on action.
Architecture hyperparameters, optimizer settings, and training schedules for LEIA and all baselines are provided in Appendix~\ref{app:hyperparams}. All models are trained on $8\times$H100 80\,GB GPUs.

% ----------------------------------------------------------------
%                               RESULTS
% ----------------------------------------------------------------
\section{Results}
\label{sec:results}

We organize our experiments around four findings. First, on MicroPlate's lattice regime (meshes up to 442,000 nodes), direct Cauchy stress supervision outperforms all indirect alternatives, and combining it with pushforward dynamics yields the best displacement and stress accuracy (Section~\ref{sec:microplate}). Second, on MicroPlate's viscoelastic regime where stress depends on deformation history through internal degrees of freedom, the stress head resolves a universal failure across four baseline architectures (Section~\ref{sec:plate_benchmarks}). Third, surrogate-guided design search evaluates 553 candidates in 30 minutes (per-sample evaluation $\sim$100--300$\times$ faster than FEM), with the improvement of the optimization goal validated by FEM (Section~\ref{sec:design_search}). Fourth, a learned confidence head predicts surrogate reliability on out-of-distribution topologies without FEM (Section~\ref{sec:ood_detection}).

% ------------------------------------------------------------------
\subsection{Accurate stress rollout on large architected meshes}
\label{sec:microplate}

\begin{figure}[h!]
  \centering
  \includegraphics[width=\textwidth]{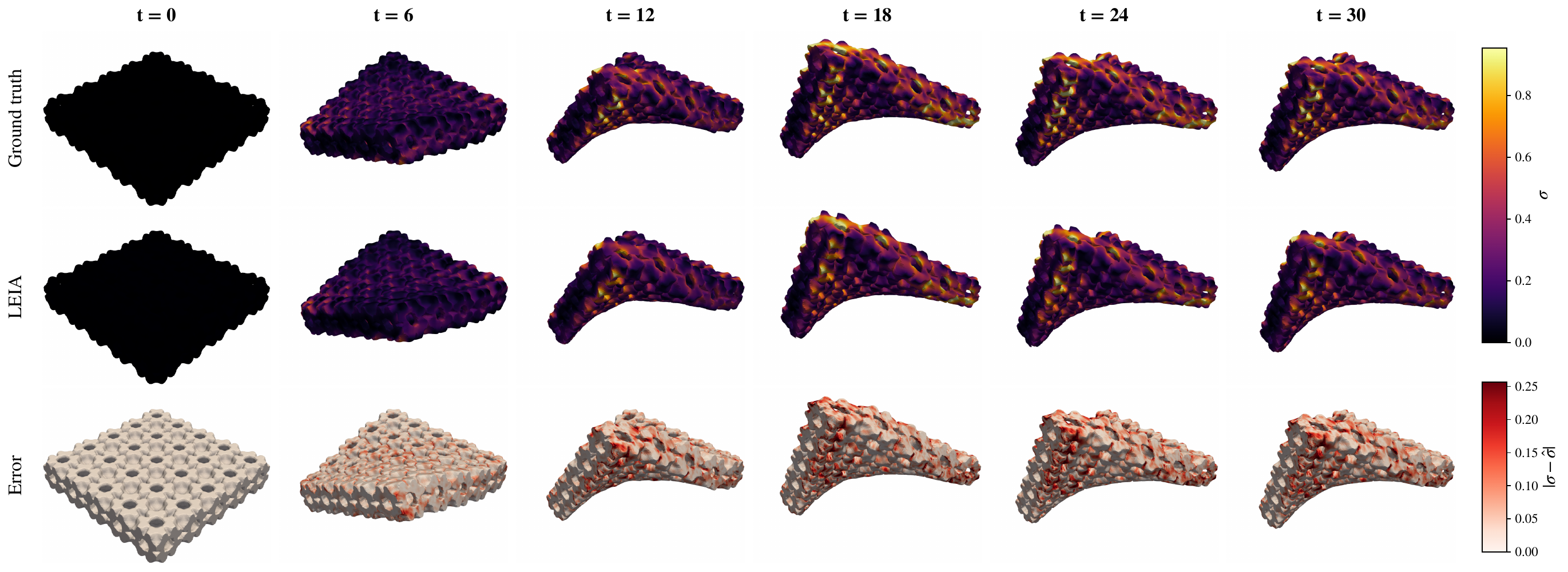}
  \caption{\footnotesize \textbf{Autoregressive rollout on a MicroPlate lattice (301,565 nodes, 1,382,904 tetrahedra).} Columns show selected timesteps. Top: ground-truth von Mises stress. Middle: LEIA prediction. Bottom: absolute von Mises stress error.}
  \label{fig:microplate_fields}
\end{figure}

We evaluate LEIA on MicroPlate's lattice regime: 63 architected lattices with meshes up to 442,000 nodes. We train on 55 lattices and hold out 8 for evaluation on unseen geometries. Fig.~\ref{fig:microplate_fields} shows a representative rollout on an unseen trajectory of a training topology. Fig.~\ref{fig:fd_curves} shows the work vs time.

We evaluate the full pipeline in two modes: \emph{teacher forcing} (ground-truth state at each step, predict one step ahead) and \emph{autoregressive rollout} (initial state only, roll out the full trajectory with user-specified boundary conditions). We report two primary metrics: (1) median per-frame displacement relative $L^2$ error, $\|\hat{\mathbf{u}} - \mathbf{u}\| / \|\mathbf{u}\|$, and (2) median per-frame Pearson correlation between predicted and ground-truth von Mises stress fields. Both are computed per trajectory and summarized by the median across test trajectories. For models without a stress head, stress is computed post-hoc from the predicted displacement via tet shape function differentiation and the constitutive law; for models with the stress head, von Mises stress is computed directly from the predicted $\hat{\bm{\sigma}}_\mathrm{sym}$. Full details of the stress calculation procedure are in Appendix~\ref{app:stress_eval}.

\begin{table}[tb]
  \caption{\footnotesize Ablation on stress supervision strategy for the LEIA tokenizer on MicroPlate, progressing from no gradient supervision to direct stress supervision. 
  The Overhead column reports additional tokenizer training time per optimizer step at fixed effective batch (128 across 8 H100s) versus the recon-only baseline. All metrics are medians on autoregressive rollout.
  }
  \label{tab:ablation_method}
  \centering
  \footnotesize
  \begin{tabular}{@{}l cccc c@{}}
    \toprule
    & \multicolumn{2}{c}{In-distribution} & \multicolumn{2}{c}{Held-out topology} & \\
    \cmidrule(lr){2-3} \cmidrule(lr){4-5}
    & Disp.\ $L^2$\,(\%)$\downarrow$ & VM corr.$\uparrow$ & Disp.\ $L^2$\,(\%)$\downarrow$ & VM corr.$\uparrow$ & Overhead \\
    \midrule
    No gradient supervision (tet eval)        & 10.83 & 0.244 & 14.60 & 0.210 & baseline \\
    Tet Sobolev                               & 7.08 & 0.410 & 7.89 & 0.388 & +53\% \\
    Autograd Sobolev                          & 9.31 & 0.327 & 12.55 & 0.252 & +1160\% \\
    Gradient head                             & 17.80 & 0.854 & 21.93 & 0.667 & +13\% \\
    Stress head                               & 13.21 & 0.870 & 17.92 & 0.654 & +6\% \\
    \midrule
    Stress head + pushforward                 & 4.68 & 0.942 & 5.39 & 0.662 & +6\% \\
    \bottomrule
  \end{tabular}
\end{table}

We compare five strategies for predicting stress, progressing from no gradient supervision to direct Cauchy stress supervision. (1)~No gradient supervision: the tokenizer is trained on displacement reconstruction only; stress is evaluated post-hoc via tet shape function differentiation of the predicted displacement $\hat{\mathbf{u}}$. (2)~Tet Sobolev: the displacement gradient is computed via tet shape function differentiation, applied to a subset of tets per training step (a deterministic furthest-point-sampled core plus a random tail, volume-weighted MSE against the per-tet ground-truth gradient). (3)~Autograd Sobolev: the displacement Jacobian $\nabla\hat{\mathbf{u}}$ is supervised via automatic differentiation through the decoder, evaluated at the FPS-subsampled query points; this adds roughly $12\times$ training time per step. (4)~Gradient head: a learned linear projection from decoder hidden states to $\nabla\mathbf{u}$ components, supervised with precomputed ground-truth gradients, avoiding the autograd cost (+13\% overhead). (5)~Stress head: a linear projection to six Cauchy stress components, supervised directly with FEM Cauchy stress, with only +6\% overhead.

Without any gradient supervision, post-hoc tet-shape-function differentiation of the predicted displacement yields VM (Von Mises stress) correlation of $0.24$ in-distribution and $0.21$ on held-out shapes. Supervising the gradient via tet shape functions provides modest gains (VM $0.41$ in-dist, $0.39$ held-out); autograd through the decoder gives smaller gains (VM $0.33$ in-dist, $0.25$ held-out), at $+1160\%$ per-step cost. The two learned-head approaches reach a substantially different range: the gradient head attains VM $0.85$ in-dist and $0.67$ held-out at $+13\%$ overhead, and the stress head attains $0.87$ and $0.65$ at $+6\%$. Adding pushforward rollout and velocity-input dynamics on top of the stress head yields the lowest displacement error in both splits ($4.68\%$ in-dist, $5.39\%$ held-out) and raises in-distribution VM correlation to $0.94$, with comparable held-out VM ($0.66$).

% ------------------------------------------------------------------
\subsection{Accurate stress rollout under homogenized microstructural change}
\label{sec:plate_benchmarks}

\begin{wraptable}{r}{0.64\textwidth}
  \vspace{-10pt}
  \caption{\footnotesize \textbf{Stress-prediction strategies on MicroPlate's load-reverse visco-hyperelastic plate.} 3 visco-hyperelastic branches, 100-step load-then-reverse protocol. 
  Both the teacher-forcing and AR (autoregressive) columns report without\,/\,with the stress head.
  Medians across 100 test trajectories. The Space column indicates whether the model operates directly on mesh nodes (phys.) or on a compressed latent representation (latent).}
  \label{tab:benchmark}
  \centering
  \footnotesize
  \begin{tabular}{@{}ll cc cc@{}}
    \toprule
    & & \multicolumn{2}{c}{Teacher forcing} & \multicolumn{2}{c}{AR rollout (100 steps)} \\
    \cmidrule(lr){3-4} \cmidrule(lr){5-6}
    Model & Space & Disp.\,(\%)$\downarrow$ & VM$\uparrow$ & Disp.\,(\%)$\downarrow$ & VM$\uparrow$ \\
    \midrule
    MGN        & phys.  & 3.88\,/\,3.52  & $-.009$\,/\,.987 & 241\,/\,9.94   & .012\,/\,.225 \\
    UPT        & latent & 13.0\,/\,6.09  & .257\,/\,.951    & 80.8\,/\,31.8  & .197\,/\,.882 \\
    PointNet   & phys.  & 21.8\,/\,3.25  & .091\,/\,.943    & 22.8\,/\,2.87  & .092\,/\,.938 \\
    Transolver & phys.  & 33.0\,/\,2.84  & .108\,/\,.941    & 40.3\,/\,2.42  & .109\,/\,.947 \\
    \midrule
    LEIA       & latent & \textbf{2.84} & .970 & 5.99 & \textbf{.986} \\
    \bottomrule
  \end{tabular}
\end{wraptable}

We now test the implicit-microstructure regime, a homogenized representation in which microstructural change is encoded as internal degrees of freedom on a continuum plate. On MicroPlate's visco-hyperelastic plate (3 visco-hyperelastic branches, 100-step load-then-reverse protocol; Appendix~\ref{app:constitutive}), stress depends on the current deformation and on 18 internal variables carried by the three branches. A model can therefore predict stress in two ways: recover it from the predicted internal variables through the constitutive law, or predict the six independent Cauchy stress components directly with the stress head. Table~\ref{tab:benchmark} reports both strategies for the four baselines. With indirect prediction, von Mises correlation stays at or below 0.26 for every baseline, across message-passing (MGN), transformer (UPT, Transolver), and set-encoder (PointNet) architectures. With the direct stress head, all four reach von Mises correlation above 0.94, and displacement accuracy also improves. Under 100-step autoregressive rollout, stress stays correlated above 0.88 for UPT and the per-frame models (PointNet, Transolver), while MGN's stress correlation drops to 0.225 alongside its displacement collapse. PointNet and Transolver are per-frame predictors with no temporal state; their rollout re-applies the model at each step using ground-truth cumulative boundary conditions, so it matches teacher forcing by construction. Among the three temporal models, LEIA degrades least: AR displacement rises from 2.84\% to 5.99\%, and AR von Mises correlation holds at 0.986, the highest value in the AR column.

% ------------------------------------------------------------------
\subsection{Surrogate-guided design search}
\label{sec:design_search}

\begin{figure}[h!]
  \centering
  \includegraphics[width=0.95\textwidth]{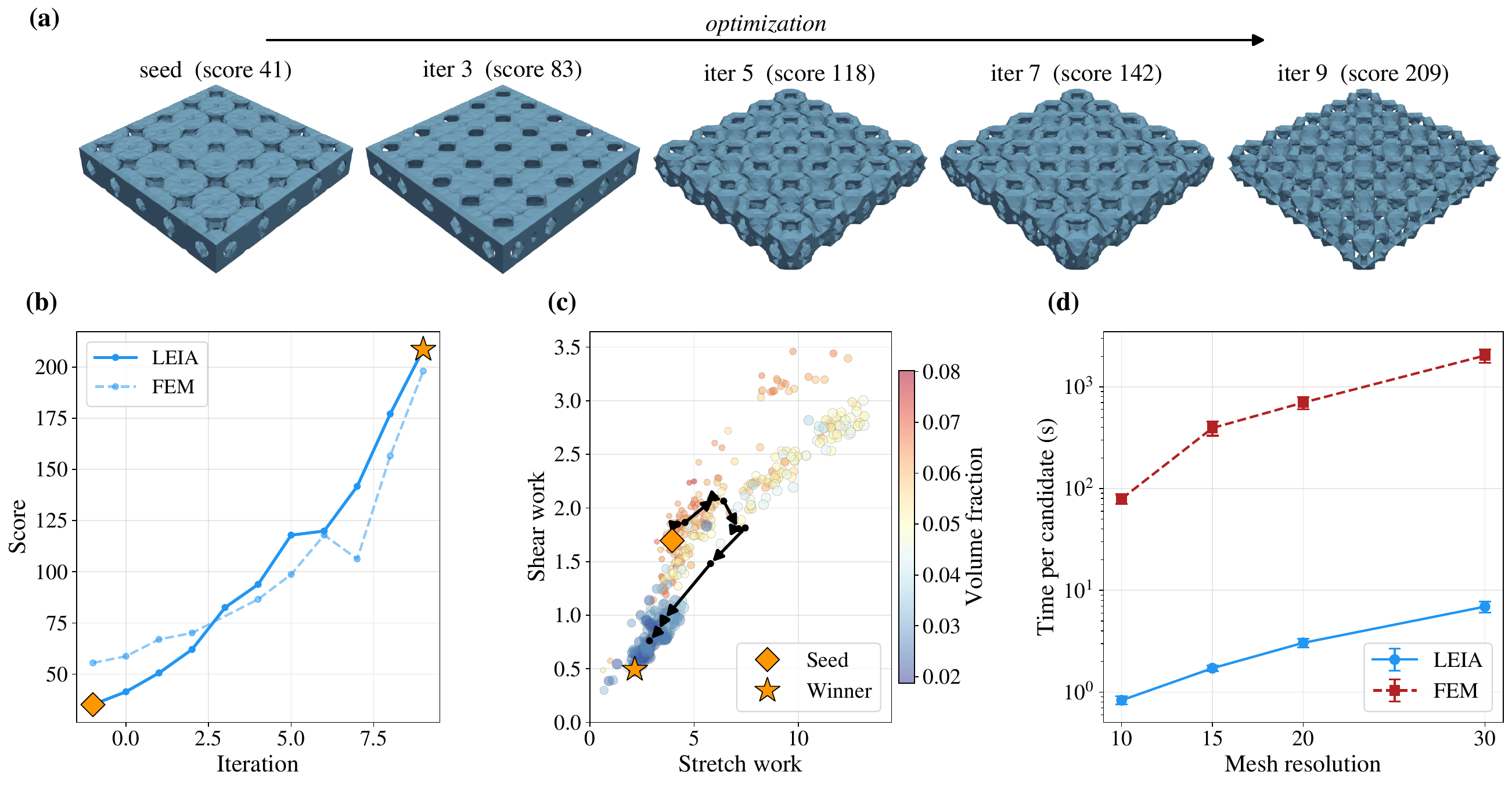}
  \caption{\footnotesize \textbf{Surrogate-guided beam search over MicroPlate.}
    \textbf{(a)}~Topology evolution: best design at selected iterations, showing progressive material removal as the search optimizes for high stretch resistance, low shear resistance, and low volume fraction.
    \textbf{(b)}~Convergence of the design metric. Solid: LEIA. Dashed: FEM validation at each iteration, confirming that the surrogate ranking is reliable.
    \textbf{(c)}~Search landscape: all evaluated candidates in stretch work vs.\ shear work space, colored by volume fraction. Arrows trace the mean search trajectory across iterations. Diamond: seed; star: winner.
    \textbf{(d)}~Per-candidate evaluation cost. LEIA (blue) and FEM (red) measured at four mesh resolutions. Error bars: standard deviation across the 11 best-per-iteration candidates at each mesh resolution.}
  \label{fig:design_search}
\end{figure}

The practical value of a fast, stress-accurate surrogate extends beyond prediction to \emph{design}: an engineer can evaluate thousands of candidate topologies with LEIA in the time FEM would evaluate a handful, then spend the FEM budget only on the most promising candidates. 

To demonstrate the practical value of stress-accurate surrogates, we use LEIA to evaluate designs in a beam search over architected-lattice mutations. The design metric is $s = W_{\text{stretch}} / (W_{\text{shear}} \cdot v_f + \varepsilon)$ with $\varepsilon = 10^{-8}$ for numerical stability, where $W_{\text{stretch}}$ and $W_{\text{shear}}$ are the work (area under the force-displacement curve) for stretch and shear loading, and $v_f$ is the normalized volume fraction. This metric favors lightweight structures that resist axial loads while remaining compliant in transverse directions. Each architected lattice's unit cell is defined by a graph of nodes and beams with associated radii. We define six mutation operators that modify the graph topology and geometry (Table~\ref{tab:mutations} in the Appendix). Each mutation is a random composition of 1--3 operators; invalid mutations are discarded and resampled (Appendix~\ref{app:mutations}). At each iteration, the top-$B$ candidates are each expanded by $M$ mutations; all new designs are meshed, evaluated with LEIA to obtain predicted force-displacement curves, and the top-$B$ are retained. In total we evaluate 553 candidate designs across 10 iterations (hyperparameters in Appendix~\ref{app:mutations}).

The search evaluates 553 candidate designs in 30 minutes of total wall-clock time. Fig.~\ref{fig:design_search}(d) shows the per-candidate LEIA/FEM speedup remains in the $\sim$100--300$\times$ range across mesh resolutions $r \in \{10, 15, 20, 30\}$. Starting from a single seed topology (\texttt{mg\_046}) with volume fraction 6.6\%, the search discovers a design with volume fraction 2.1\% and 10 struts (down from 15) that achieves a $3.6\times$ improvement in the design metric as validated by FEM ground truth. FEM validation at every iteration confirms that the surrogate's ranking tracks the ground truth throughout the search (Fig.~\ref{fig:design_search}).

\subsection{Out-of-distribution design generation and surrogate confidence}
\label{sec:ood_detection}

\begin{figure}[h!]
  \centering
  \includegraphics[width=\textwidth]{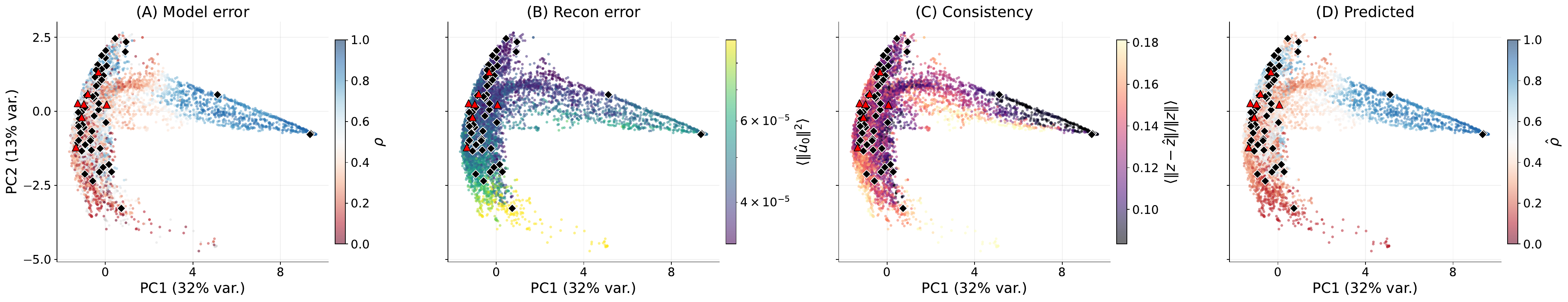}
  \caption{\footnotesize \textbf{OOD-detection landscape.}
    PCA of the tokenizer's mean-pooled latent fit on the training plates and applied to mutated candidates.
    Black diamonds: training plates. Red triangles: held-out plates. Light grey dots in (A): candidates whose FEM diverged.
    \textbf{(A)} FEM ground-truth model error: $\rho$ is the Pearson correlation between LEIA and FEM-derived von Mises stress fields.
    \textbf{(B)} Tokenizer round-trip error $\langle\|\hat{\mathbf{u}}_0\|^2\rangle$ with $\hat{\mathbf{u}}_0 = \mathrm{decode}(\mathrm{encode}(\mathbf{0}))$.
    \textbf{(C)} Encoder-cycle inconsistency $\langle\|\mathbf{z}-\hat{\mathbf{z}}\|/\|\mathbf{z}\|\rangle$ with $\hat{\mathbf{z}}_t = \mathrm{encode}(\mathrm{decode}(\mathbf{z}_t))$.
    \textbf{(D)} Predicted $\hat{\rho}$ from a learned confidence head.}
  \label{fig:ood_landscape}
\end{figure}

Surrogate-guided exploration could produce shapes far from the training distribution, where LEIA's predictions may be unreliable. Detecting when the surrogate enters this regime, without running FEM (which defeats the purpose of the surrogate), is essential for a practical design loop. To test this, we generate a diverse set of 7{,}981 mutated MicroPlate candidates using random walks and novelty search that deliberately encourage out-of-distribution exploration (details in Appendix~\ref{app:ood_detection}). Fig.~\ref{fig:ood_landscape} projects these candidates onto the tokenizer's mean-pooled latent PCA fit on the 55 training plates.

To predict surrogate reliability without running FEM, we evaluate two FEM-free signals: the tokenizer's frame-0 round-trip reconstruction error and an encoder-cycle inconsistency that measures whether dynamics-predicted latents survive a decode-then-re-encode round trip. We also train a learned confidence head that combines these signals with the tokenizer's mean-pooled latent and 15-dimensional graph statistics to predict LEIA's per-plate von Mises correlation $\rho$. Signal definitions, head architecture, and training protocol are in Appendix~\ref{app:ood_detection}.

The tokenizer's frame-0 reconstruction error (Fig.~\ref{fig:ood_landscape}B) flags the lower-left region of the latent cloud where geometries are far from the training distribution, but is otherwise uncorrelated with FEM-derived $\rho$ globally. The encoder-cycle inconsistency over a 10-step uniaxial-stretch rollout (Fig.~\ref{fig:ood_landscape}C) identifies a different failure mode in the upper region, where rolled-out latents fail to round-trip cleanly through the encoder. Both signals miss the central stripe of the latent cloud where FEM-derived $\rho$ is low despite low reconstruction error and low inconsistency; the dynamics model produces configurations that are wrong but still within the distribution recognizable by the tokenizer, so round-trip checks cannot catch the error. 

To close this gap, we train a small learned head on the 6{,}919 FEM-validated candidates. As shown in Fig.~\ref{fig:ood_landscape}(D), the learned head captures the spatial pattern of FEM-derived $\rho$ across the latent cloud, including the central stripe that the free signals miss. Free signals filter obvious geometric outliers cheaply; the learned head is needed for cases that are in-distribution but inaccurate. Together they enable a surrogate-assisted design loop: generate many candidates with mutation-based search, screen with the learned head to rank candidates and flag uncertain ones, and reserve FEM for validating the most promising designs. The confidence signal could also guide active data collection for out-of-distribution topologies, enabling targeted model fine-tuning in future work.

\section{Discussion}\label{sec:discussion}

\paragraph{An interactive environment for engineering design.}
LEIA provides a real-time interactive loop for architected material design: the user specifies boundary conditions, observes the resulting deformation and stress in real time, and iterates. This enables workflows that are impractical with finite element simulation alone: rapid screening of candidate topologies for target properties, interactive exploration of loading paths, and surrogate-guided search over large design spaces. 
LEIA can also decode at different resolutions, which could enable fast coarse rendering for exploration and zooming in to finer detail for inspection, a practical feature for interactive engineering workflows. 
A current limitation is geometric scope: LEIA is evaluated only on cubically symmetric lattice plates with a fixed $5\times5\times1$ tiling under a 4-degree-of-freedom action space; transfer to arbitrary three-dimensional geometries, non-lattice microstructures, or additional action space is untested. Extending LEIA to these regimes is a natural direction for future work.

\paragraph{Surrogate-assisted design and active learning.}
Section~\ref{sec:design_search} demonstrates surrogate-guided beam search as one instantiation of LEIA's role as a design tool. Several extensions are natural.
\emph{Active learning}: the learned confidence head of Section~\ref{sec:ood_detection} can provide a cheap signal for identifying topologies where the surrogate is unreliable; such topologies would be candidates for new FEM data collection, enabling targeted dataset growth that concentrates compute where it is most needed. A limitation is that the current system performs no active learning at inference time; an interactive session could queue on-the-fly FEM validation when the user navigates into regions of boundary-condition space where LEIA is least confident, closing the surrogate-improvement loop during use. The beam search already demonstrates the core loop (surrogate screen $\to$ FEM validate), and active learning would extend it further.
\emph{Learned policies}: For future work, it would also be interesting to test neural policies trained against LEIA as a differentiable environment, selecting actions to match engineering objectives, replacing beam-search controller of Section~\ref{sec:design_search} with a fully learned closed loop.

\section{Conclusion}
Architected materials offer vast design potential, but realizing it requires navigating a combinatorial space of geometries and loading conditions that finite element simulation alone cannot explore at scale. LEIA provides a real-time interactive environment for this exploration: engineers specify boundary conditions, observe deformation and stress, and iterate across candidate topologies at a pace that enables surrogate-guided design search. A learned confidence head predicts where the surrogate is reliable, enabling targeted allocation of FEM compute to the most promising or uncertain candidates. We release LEIA together with MicroPlate and baseline implementations to serve as a foundation for further work. Looking ahead, augmenting LEIA with additional physics, such as fluid-structure coupling, thermal transport, or electrochemical degradation, would broaden its applicability to multi-functional demands placed on hierarchical materials in practice.

\clearpage
\bibliographystyle{unsrtnat}
\bibliography{references}

%----------------------------------------------------
%                       APPENDIX
%----------------------------------------------------
\clearpage
\appendix

\section{Constitutive Models and FEM Solver}
\label{app:constitutive}

We consider finite deformation solid mechanics on a body in reference configuration $\Omega_0$. The displacement field $\mathbf{u}(\mathbf{X}, t)$ maps material points $\mathbf{X}$ to deformed positions $\mathbf{x} = \mathbf{X} + \mathbf{u}$, with deformation gradient $\mathbf{F} = \mathbf{I} + \nabla\mathbf{u}$, Jacobian $J = \det\mathbf{F}$, and right Cauchy-Green tensor $\mathbf{C} = \mathbf{F}^\top\mathbf{F}$. The balance of linear momentum in the reference configuration reads
\begin{equation}
  \mathrm{Div}\,\mathbf{P} = \rho_0\,\ddot{\mathbf{u}},
  \tag{A1}
\end{equation}
where $\mathbf{P}$ is the first Piola-Kirchhoff stress and $\rho_0$ is the reference density. For hyperelastic materials the stress derives from a free energy density $\Psi_\mathrm{R}$ per unit reference volume:
\begin{equation}
  \mathbf{P} = 2\mathbf{F}\,\frac{\partial\Psi_\mathrm{R}}{\partial\mathbf{C}}.
  \tag{A2}
\end{equation}
The Cauchy stress is $\bm{\sigma} = J^{-1}\mathbf{P}\mathbf{F}^\top$.

\paragraph{Neo-Hookean hyperelasticity.}
A compressible Neo-Hookean free energy is used,
\begin{equation}
  \Psi_\mathrm{R} = \frac{\mu}{2}(\mathrm{tr}\,\mathbf{C} - 3) - \mu\ln J + \frac{\lambda}{2}(\ln J)^2,
  \tag{A3}
\end{equation}
where the parameters $\mu = 1$, $\lambda = 10$. In the Neo-Hookean setup we solve the systems in the quasi-static limit: the inertial term vanishes ($\rho_0\ddot{\mathbf{u}} \to \mathbf{0}$), and there are no additional internal variables.

\paragraph{Finite viscoelasticity with inertia.}
In contrast, the visco-hyperelastic plate has both inertia and rate-dependent internal state, using the constitutive model of Stewart and Anand~\cite{stewart2024viscoelasticity}. The free energy decomposes into equilibrium and non-equilibrium contributions:
\begin{equation}
  \Psi_\mathrm{R} = \Psi_\mathrm{R}^\mathrm{eq}(\bar{\mathbf{C}}) + \sum_{i=1}^{N_\mathrm{b}}\Psi_\mathrm{R}^{\mathrm{neq}(i)}(\bar{\mathbf{C}}, \mathbf{A}^{(i)}) + \frac{\kappa}{2}(J-1)^2,
  \tag{A4}
\end{equation}
where $\bar{\mathbf{C}} = J^{-2/3}\mathbf{C}$ is the distortional Cauchy-Green tensor and $\kappa$ is the bulk modulus enforcing near-incompressibility. The equilibrium term uses the Arruda-Boyce eight-chain model:
\begin{equation}
  \Psi_\mathrm{R}^\mathrm{eq}(\bar{\mathbf{C}}) = G_\mathrm{eq}\,\lambda_L^2\left[\frac{\bar{\lambda}}{\lambda_L}\,\beta_L\!\left(\frac{\bar{\lambda}}{\lambda_L}\right) + \ln\frac{\beta_L(\bar{\lambda}/\lambda_L)}{\sinh\beta_L(\bar{\lambda}/\lambda_L)} - C_0\right],
  \tag{A5}
\end{equation}
where $\bar{\lambda} = \sqrt{\mathrm{tr}(\bar{\mathbf{C}})/3}$ is the chain stretch, $\lambda_L$ is the locking stretch, $\beta_L(z) = z(3 - z^2)/(1 - z^2)$ is an approximation of the inverse Langevin function, and $C_0$ is a constant ensuring $\Psi_\mathrm{R}^\mathrm{eq} = 0$ in the reference configuration. Each non-equilibrium branch uses a neo-Hookean form in the distortional elastic deformation relative to the internal variable $\mathbf{A}^{(i)}$:
\begin{equation}
  \Psi_\mathrm{R}^{\mathrm{neq}(i)}(\bar{\mathbf{C}}, \mathbf{A}^{(i)}) = \frac{G_\mathrm{neq}^{(i)}}{2}\bigl(\bar{\mathbf{C}}:(\mathbf{A}^{(i)})^{-1} - 3\bigr).
  \tag{A6}
\end{equation}
Each internal variable evolves toward thermodynamic equilibrium:
\begin{equation}
  \dot{\mathbf{A}}^{(i)} = \frac{1}{\tau^{(i)}}\bigl(\bar{\mathbf{C}} - \mathbf{A}^{(i)}\bigr), \quad i = 1, \ldots, N_\mathrm{b}, \quad \det\mathbf{A}^{(i)} = 1,
  \tag{A7}
\end{equation}
with relaxation times $\tau^{(i)}$. Material parameters: $G_\mathrm{eq} = 200$\,kPa, $\lambda_L = 10$, $\kappa = 400{,}000$\,kPa, $\rho_0 = 1.3\times 10^{-5}$\,kg/mm$^3$, $N_\mathrm{b} = 3$ branches:

\begin{center}
\footnotesize
\begin{tabular}{@{}lccc@{}}
\toprule
Branch & 1 & 2 & 3 \\
\midrule
$G_\mathrm{neq}$ (kPa) & 300 & 600 & 150 \\
$\tau$ (s) & $0.001$ & $0.2$ & $3.0$ \\
\bottomrule
\end{tabular}
\end{center}

\paragraph{Initial and boundary conditions.}
All simulations start from the undeformed reference configuration: $\mathbf{u} = \mathbf{0}$, and for the visco-hyperelastic plate additionally $\dot{\mathbf{u}} = \mathbf{0}$, $\ddot{\mathbf{u}} = \mathbf{0}$, $\mathbf{A}^{(i)} = \mathbf{I}$ (stress-free equilibrium). The left face ($x = x_\mathrm{min}$) is fully clamped (zero displacement); the right face ($x = x_\mathrm{max}$) is the controlled grip. The remaining faces are traction-free. 

\paragraph{FEM solver details.}
Both regimes are solved via the standard Galerkin weak form of the momentum balance. For the visco-hyperelastic plate, the mixed displacement-pressure formulation finds $(\mathbf{u}, p) \in V_h \times Q_h$ such that
\begin{equation}
  \int_{\Omega_0} \mathbf{P} : \nabla\mathbf{v}\,\mathrm{d}V + \int_{\Omega_0} \rho_0\,\ddot{\mathbf{u}} \cdot \mathbf{v}\,\mathrm{d}V = 0 \quad \forall\,\mathbf{v} \in V_h, \qquad
  \int_{\Omega_0} \left(J - 1 + \frac{p}{\kappa}\right) q\,\mathrm{d}V = 0 \quad \forall\,q \in Q_h,
  \tag{A8}
\end{equation}
where the first equation is the momentum balance and the second enforces near-incompressibility through the pressure penalty. For the lattice regime, the quasi-static pure displacement formulation finds $\mathbf{u} \in V_h$ such that
\begin{equation}
  \int_{\Omega_0} \mathbf{P} : \nabla\mathbf{v}\,\mathrm{d}V = 0 \quad \forall\,\mathbf{v} \in V_h.
  \tag{A9}
\end{equation}

The visco-hyperelastic plate was generated with FEniCSx (DOLFINx 0.8) using Taylor-Hood elements: CG2 (quadratic Lagrange) for displacement and CG1 (linear Lagrange) for pressure. Internal variables $\mathbf{A}^{(i)}$ are stored at quadrature points (degree 2). Nonlinear solver: Newton iteration with MUMPS direct LU ($\mathrm{atol} = \mathrm{rtol} = 10^{-8}$ on the displacement increment norm, max 50 iterations). Time integration: Newmark-beta (trapezoidal rule, $\gamma = 0.5$, $\beta = 0.25$, $\mathrm{d}t = 0.002$\,s, 5 FEM substeps per action step).

The lattice regime dataset was generated with legacy FEniCS (\texttt{dolfin} 2019.1.0) using CG1 (linear Lagrange) displacement elements, pure displacement formulation. Newton iteration with MUMPS ($\mathrm{rtol} = 10^{-3}$, $\mathrm{atol} = 10^{-10}$ on the residual norm, max 15 iterations). Quasi-static incremental loading over 30 steps. Actions use the same per-DOF block sampling as the visco-hyperelastic plate, with per-step magnitudes $0.15$ (stretch, shear$_y$, shear$_z$) and $0.08$ (twist). FEM data generation ran on CPU, with up to 200 parallel FEniCS workers (each trajectory solved on a single worker).

\paragraph{Stress calculation from predicted displacement.}
\label{app:stress_eval}
For models without a stress head, we compute stress from the predicted displacement field using the standard finite element procedure. Shape function gradients on the tetrahedral mesh yield the deformation gradient $\mathbf{F} = \mathbf{I} + \nabla\hat{\mathbf{u}}$ at element centroids, the constitutive law maps $\mathbf{F}$ to Cauchy stress $\bm{\sigma}$, and volume-weighted averaging projects element stresses to nodes. The von Mises stress is $\sigma_\mathrm{VM} = \sqrt{\tfrac{3}{2}\,\mathbf{s}:\mathbf{s}}$, where $\mathbf{s} = \bm{\sigma} - \tfrac{1}{3}(\mathrm{tr}\,\bm{\sigma})\,\mathbf{I}$ is the deviatoric stress. For models with the stress head, von Mises stress is computed directly from the predicted $\hat{\bm{\sigma}}_\mathrm{sym}$ without any mesh-based differentiation. For ablation variants with a gradient head, the predicted displacement gradient is mapped to Cauchy stress via the constitutive law.

\section{MicroPlate Benchmark Details}
\label{app:microplate}

\paragraph{Graph representation.}
The unit cell of each architected lattice is defined by a compact graph $\mathcal{G} = (\mathcal{V}, \mathcal{E})$, where $\mathcal{V} = \{\mathbf{v}_i \in \mathbb{R}^3\}$ are node positions and $\mathcal{E} = \{(i, j, r)\}$ are beams connecting pairs of nodes with radius $r$.

An example graph with 4 nodes and 4 beams:
\begin{center}
\footnotesize
\begin{verbatim}
{"nodes": [[0.24, 0.61, 0.46],
           [0.15, 0.18, 0.19],
           [0.44, 0.57, 0.85],
           [0.64, 0.12, 0.51]],
 "beams": [{"idx": [0,1], "r": 0.017},
           {"idx": [1,2], "r": 0.011},
           {"idx": [2,3], "r": 0.016},
           {"idx": [3,0], "r": 0.014}]}
\end{verbatim}
\end{center}

\begin{figure}[h!]
  \centering
  \includegraphics[width=0.45\textwidth]{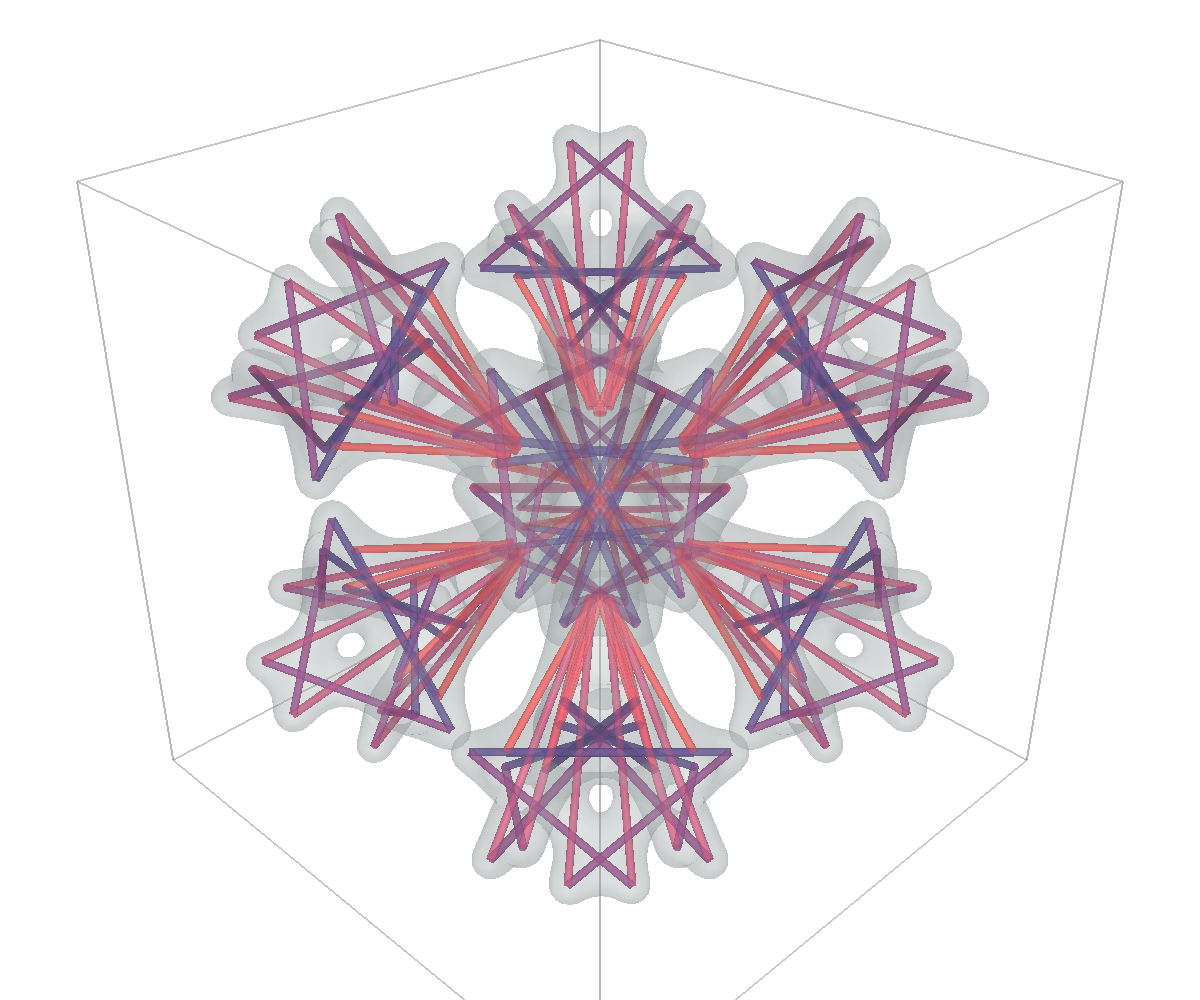}
  \caption{\footnotesize \textbf{Unit cell of a MicroPlate lattice.} The compact graph (4 beams, shown as colored rods) is replicated under all 48 operations of the cubic symmetry group $O_h$, producing $48 \times 4 = 192$ beams whose smooth-min-blended signed distance field yields the final mesh (gray). Each color represents one seed beam and all its symmetric copies.}
  \label{fig:rve_graph}
\end{figure}

\paragraph{Cubic symmetry.}
To ensure that the resulting microstructure has full cubic symmetry, every beam in the graph is replicated under all 48 operations of the octahedral symmetry group $O_h$ (the 6 coordinate permutations combined with the 8 sign combinations). A graph with $|\mathcal{E}|$ beams thus produces $48\,|\mathcal{E}|$ beams in the expanded unit cell (Fig.~\ref{fig:rve_graph}). This ensures that the mechanical response of each topology is invariant under the full cubic point group.

\paragraph{Mesh generation.}
The expanded beam geometry is evaluated as a signed distance field (SDF) on a regular voxel grid at 30 voxels per unit cell, using smooth-min blending between beam capsules to ensure connected junctions. The unit cell is tiled $5\times5\times1$ into a $[10\times10\times2]$ plate geometry via modular coordinate wrapping: for each evaluation point, the SDF kernel maps the global plate coordinate back to the local unit cell coordinate, so the lattice pattern repeats seamlessly across tile boundaries without explicit stitching. An auto-scaling step ensures the structure fills the unit cell boundary, and a percolation check verifies that the tiled lattice forms a single connected component. The isosurface is extracted via marching cubes, producing a triangular surface mesh that is then volumetrically meshed with TetGen. Mesh sizes range from 71,000 to 442,000 nodes (250,000--2,020,000 tetrahedra). The complex stress concentration patterns at strut junctions distinguish the lattice plates from the homogeneous plate, where stress fields are smooth.

\begin{figure}[h!]
  \centering
  \includegraphics[width=\textwidth]{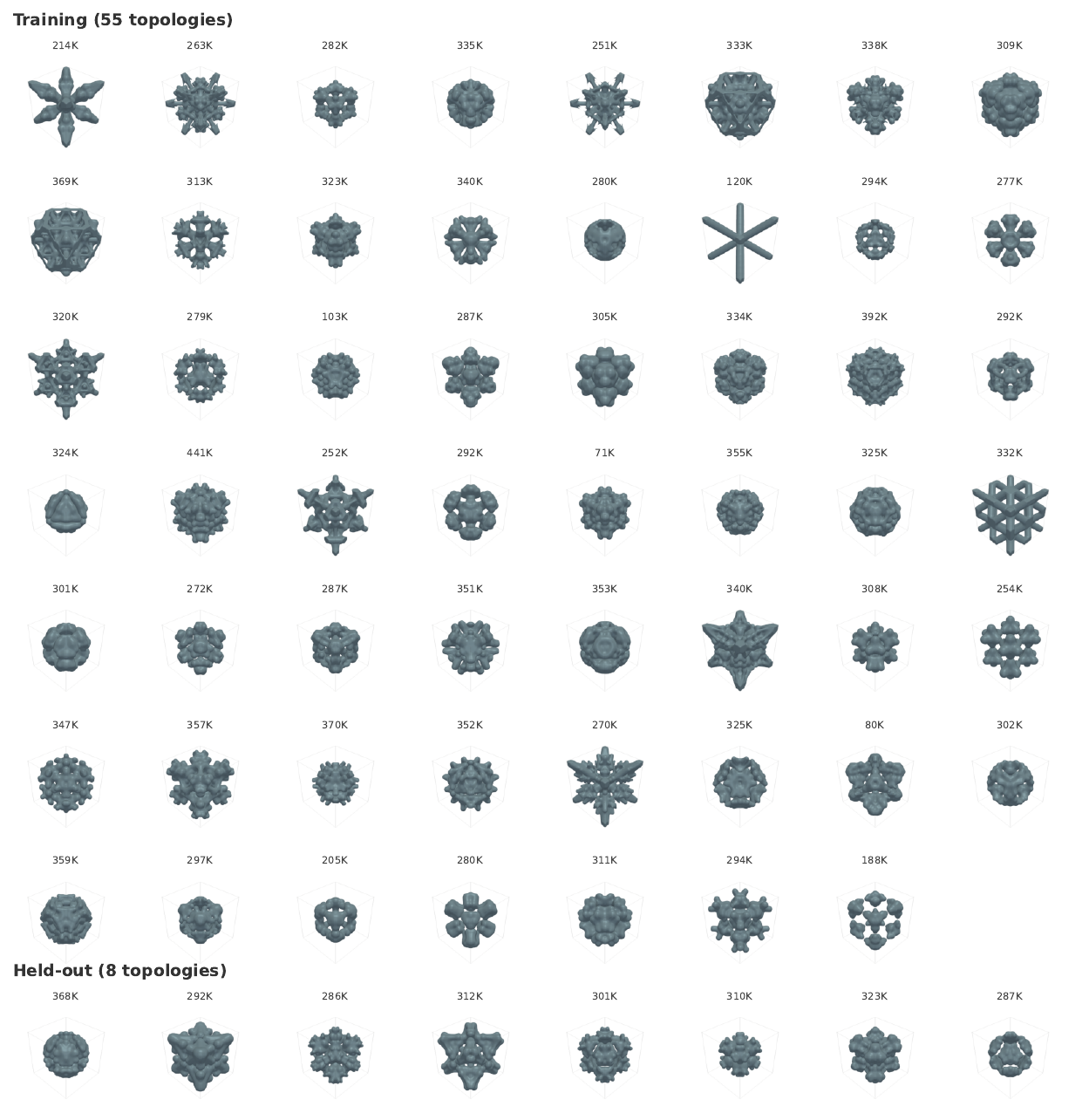}
  \caption{\footnotesize \textbf{Library of MicroPlate unit cells.} Unit cells (RVEs) corresponding to the 63 architected lattices, shown after $O_h$ symmetry expansion. Top: 55 training topologies. Bottom: 8 held-out topologies for evaluating generalization to unseen geometries. The number above each cell is the plate mesh size (in thousands of nodes) after $5\times5\times1$ tiling and TetGen volumetric meshing.}
  \label{fig:microplate_dashboard}
\end{figure}

\begin{figure}[h!]
  \centering
  \includegraphics[width=0.85\textwidth]{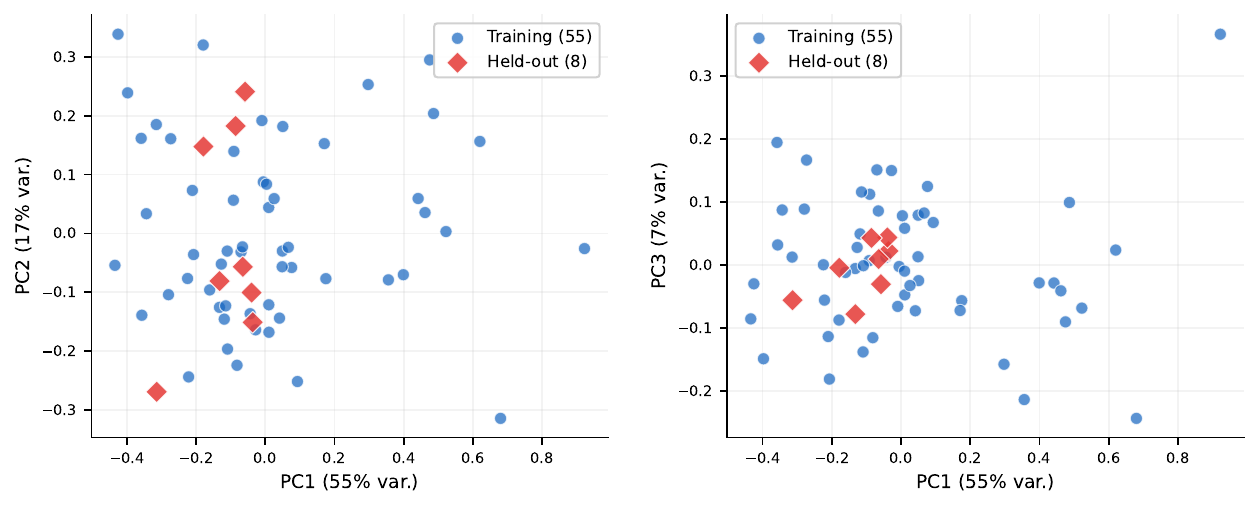}
  \caption{\footnotesize \textbf{PCA projection of architected-lattice features.} PCA projection of the 15-dimensional graph feature vectors (node/beam counts, spatial spread, beam lengths, radii, connectivity) for all 63 architected lattices.}
  \label{fig:microplate_pca}
\end{figure}

\paragraph{Dataset statistics.}
Each topology has multiple boundary condition trajectories (8,836 total across 63 topologies), each 30 frames under 4-DOF random discrete actions simulated using different random seeds. Material: Neo-Hookean, $\mu = 1$, $\lambda = 10$. MicroPlate is split by topology into 55 training and 8 held-out plates. Each of the 55 training plates have more than 100 FEM trajectories, the first 100 trajectories (total 5500) are used for training and the 101-th trajectory (a different random loading path) is used for evaluation.
The 1,000 visco-hyperelastic-plate trajectories are split 800/100/100 (train/val/test) with seed 42. Figure~\ref{fig:microplate_dashboard} shows all 63 MicroPlate unit cells grouped by split, Fig.~\ref{fig:microplate_pca} shows the PCA distribution of topology features, and Table~\ref{tab:benchmarks} compares MicroPlate's scale against existing solid mechanics benchmarks.

\begin{table}[h!]
  \caption{\footnotesize \textbf{MicroPlate vs.\ existing solid mechanics benchmarks.} MicroPlate is two orders of magnitude larger in mesh size, and it is a temporal 3D benchmark with microstructure diversity.}
  \label{tab:benchmarks}
  \centering
  \small
  \begin{tabular}{@{}lrrcc@{}}
    \toprule
    Benchmark & Nodes & Temporal & 3D & Nonlinear \\
    \midrule
    Geo-FNO Elasticity~\cite{li2023geofno}         & $\sim$1K       & No  & No  & No \\
    MGN DeformingPlate~\cite{pfaff2021mgn}          & $\sim$1K     & Yes & No  & No \\
    LatticeGraphNet~\cite{jain2024latticegraphnet}  & $\sim$5K  & Yes & Yes & Yes \\
    \textbf{MicroPlate (ours)}                      & \textbf{71K--442K} & \textbf{Yes} & \textbf{Yes} & \textbf{Yes} \\
    \bottomrule
  \end{tabular}
\end{table}

\clearpage
\section{Detailed Training Procedure}
\label{app:hyperparams}

LEIA is trained in two stages. First, the tokenizer is trained on individual frames to reconstruct displacement (and, when using the stress head, Cauchy stress) from the latent representation. Second, with the tokenizer frozen, the dynamics transformer is trained to predict the next latent state from the current state and boundary condition action. 

\paragraph{Stress supervision.}
The tokenizer is trained per frame with a combined loss $\mathcal{L}_\mathrm{tok} = \|\hat{\mathbf{u}} - \mathbf{u}\|^2 + \beta\,\|\hat{\bm{\sigma}}_\mathrm{sym} - \bm{\sigma}_\mathrm{sym}^\mathrm{FEM}\|^2$, where $\bm{\sigma}_\mathrm{sym}^\mathrm{FEM}$ is the Cauchy stress from the finite element solver, stored as 6 symmetric components. Because stress magnitudes can differ from displacements by orders of magnitude, we normalize each output component to zero mean and unit variance before computing the loss. The stress head itself is a single $\mathrm{Linear}(H, 6)$ layer ($< 0.02$\% of model parameters). The same stress head applies across architectures: we attach it to UPT, Transolver, PointNet, and MeshGraphNets by replacing each model's output layer with one producing $\mathbf{u}(3) + \bm{\sigma}_\mathrm{sym}(6) = 9$ output dimensions. No other architectural changes are made.

The output dimension of the stress head is constant regardless of constitutive law complexity. The Cauchy stress $\bm{\sigma}$ is always a symmetric $3 \times 3$ tensor with 6 independent components, whether the material is Neo-Hookean or viscoelastic. By contrast, predicting all internal variables scales from 3 output dimensions (Neo-Hookean, displacement only) to 22 (3-branch visco-hyperelastic: displacement + 18 $\mathbf{A}^{(i)}$ components + pressure) or higher for more complex constitutive laws.

\paragraph{Furthest point sampling.}
For the lattice regime's meshes (up to 442,000 nodes), encoding all nodes simultaneously is prohibitive. We subsample $S$ query points via furthest point sampling (FPS), which provides approximately uniform spatial coverage of the mesh. The Perceiver encoder compresses these $S$ sampled node features into $K$ latent tokens. At decoding time, the decoder reconstructs the field at all $N$ mesh nodes by querying the latent tokens from each node's position.

\paragraph{Dynamics training.}
With the tokenizer frozen, the dynamics transformer is trained to predict the next latent state: $\mathcal{L}_\mathrm{dyn} = \|\mathbf{z}_{t+1}^\mathrm{pred} - \mathbf{z}_{t+1}^\mathrm{target}\|^2$, where $\mathbf{z}_{t+1}^\mathrm{target}$ is the encoding of the ground-truth next frame. Optionally, pushforward training \cite{brandstetter2022message} unrolls the model for $k$ steps and averages the loss over the rollout, reducing the distribution shift between teacher-forced training and autoregressive inference. For viscoelastic systems where stress depends on deformation history, the dynamics model can additionally be conditioned on the latent velocity $\mathbf{z}_t - \mathbf{z}_{t-1}$, concatenated with the current state before projection. Per-dataset training configurations are reported in Appendix~\ref{app:hyperparams}.

\paragraph{Ablations.}
The first five rows in Table~\ref{tab:ablation_method} share the same architecture (n\_input\_frames=1, K=512, H=1024, 6 encoder layers, no decoder layers), the same dynamics recipe (single-step continuous), and the same downstream pipeline; only the tokenizer's stress-supervision strategy differs. The last row applies our production dynamics enhancements (pushforward rollout and velocity-input conditioning) on top of the stress head, isolating the contribution of dynamics-side training from tokenizer supervision. The supervision-loss term and its weight differ per row. 
The stress head uses $\beta = 1$ on the standardized Cauchy MSE term from the tokenizer training loss, with linear warmup over the first 10K of 30K tokenizer steps. Tet and autograd Sobolev use Sobolev MSE on $\nabla\mathbf{u}$ with $\beta = 3$ ($\beta = 30$ destabilizes reconstruction for these paths). The gradient head uses Sobolev MSE with $\beta = 30$, the larger weight yielding the higher VM correlation reported in Table~\ref{tab:ablation_method}.

\paragraph{Baselines.}
All four baseline architectures (UPT, Transolver, PointNet, MeshGraphNets) are trained end-to-end on the same data splits. The only modification to each baseline is the output dimension: 3 (displacement only), 9 (displacement + 6 Cauchy stress components with the stress head), or 22 (displacement + 18 internal variable components + pressure, without the stress head on the visco-hyperelastic plate). 

Table~\ref{tab:hyperparams} reports the full architecture and optimizer configurations of LEIA. All models are trained on $8\times$H100 80\,GB GPUs unless noted otherwise.

\begin{table}[h!]
  \caption{\footnotesize \textbf{Hyperparameters for LEIA and all baselines.}}
  \label{tab:hyperparams}
  \centering
  \footnotesize
  \begin{tabular}{@{}p{3cm}p{6.5cm}p{4cm}@{}}
    \toprule
    Model & Architecture & Training \\
    \midrule
    LEIA tokenizer (lattice regime) & Fourier PE (16 bands), $H{=}1024$, $K{=}512$ latents, $L_\mathrm{enc}{=}6$, $L_\mathrm{dec}{=}0$, 4 heads, dropout 0.1, input: 3-frame disp(9) + Fourier PE of ref coords, FPS 8192{+}2048 random nodes $\to$ $K$ latents, $\approx$103.5M params & AdamW $10^{-4}$, wd $10^{-5}$, CosineAnnealingLR, batch 8 ($\times$2 accum), bf16, 100K steps, 8$\times$H100 \\[2pt]
    LEIA dynamics (lattice regime)  & $H{=}1024$, 12 layers, 8 heads, dropout 0.3, action: 4-dim cumulative BC, FiLM: MLP($4{\to}256{\to}256$, GELU) + per-layer SiLU proj to $(\gamma,\beta){\times}2$, velocity input ($\mathbf{z}_t{-}\mathbf{z}_{t-1}$), $\approx$205.2M params & AdamW $10^{-4}$, wd $10^{-4}$, pushforward ($k{=}10$), noise std 0.02, warmup (10 ep) + CosineAnnealingLR, batch 64, bf16, 300 ep, 8$\times$H100 \\[2pt]
    LEIA tokenizer (visco-hyperelastic plate) & $H{=}256$, $K{=}256$ latents, $L_\mathrm{enc}{=}8$, $L_\mathrm{dec}{=}4$, 4 heads, dropout 0.1, input: 3-frame disp(9) + Fourier PE, stress head, all 363 nodes (no FPS), $\approx$11M params & AdamW $3{\times}10^{-4}$, wd $10^{-5}$, CosineAnnealingLR, batch 64, bf16, 30K steps, 8$\times$H100 \\[2pt]
    LEIA dynamics (visco-hyperelastic plate)  & $H{=}256$, 12 layers, 4 heads, dropout 0.3, action: 4-dim discrete BC, velocity input ($\mathbf{z}_t{-}\mathbf{z}_{t-1}$), $\approx$13M params & AdamW $3{\times}10^{-4}$, wd $10^{-4}$, pushforward ($k{=}10$), CosineAnnealingLR, batch 64, bf16, 300 ep, 8$\times$H100 \\
    \bottomrule
  \end{tabular}
\end{table}

\section{Beam Search Mutation Operators}
\label{app:mutations}

The surrogate-guided design search (Section~\ref{sec:design_search}) uses beam search with stochastic graph mutations. At each iteration, each of the top-$B{=}5$ candidates is expanded by generating $\lfloor M \times 1.5 \rfloor{=}12$ mutation attempts per parent (over-generating to account for invalid mutations), for a target of $M{=}8$ valid candidates per parent. Each mutation randomly composes 1--3 operators (with probabilities 0.6, 0.3, 0.1 for 1, 2, 3 operators respectively), drawn from the distribution in Table~\ref{tab:mutations} and applied sequentially to the graph. After each mutation, the graph is validated: node coordinates must lie in $[-1.5, 1.5]$, beam radii in $(0, 0.5]$, volume fraction in $[0.001, 0.5]$, and the lattice must remain connected (all nodes reachable by graph traversal). Invalid mutations are rejected and resampled up to 20 times. Valid candidates are meshed via the pipeline in Appendix~\ref{app:microplate} and evaluated with LEIA on two loading directions (stretch and shear) for 20 steps each to obtain force-displacement curves. The design metric is $s = W_{\text{stretch}} / (W_{\text{shear}} \cdot v_f + \varepsilon)$, where work $W$ is the trapezoidal integral of $|\text{force}|$ over $|\text{displacement}|$, $v_f$ is the normalized volume fraction, and $\varepsilon = 10^{-8}$. The search runs for $D{=}10$ iterations, evaluating 553 total candidates from seed topology \texttt{mg\_046}. 
% The search is reproducible with random seed 42.

\begin{table}[h!]
  \caption{\footnotesize \textbf{Mutation operators for surrogate-guided architected-lattice search.} Each mutation randomly composes 1--3 operators (drawn with the listed probabilities), validated for graph connectivity before evaluation.}
  \label{tab:mutations}
  \centering
  \small
  \begin{tabular}{@{}llll@{}}
    \toprule
    Operator & Description & Parameters & Prob. \\
    \midrule
    Perturb nodes   & Displace 1--3 random nodes         & $\delta \sim \mathcal{N}(0, 0.08)$ per axis  & 25\% \\
    Scale radii     & Multiply radii of 1--4 random beams & Factor $\sim e^{\mathcal{N}(0, 0.15)}$        & 20\% \\
    Add beam        & Connect two unconnected nodes       & Radius $\sim \mathrm{Uniform}(0.005, 0.04)$  & 15\% \\
    Remove beam     & Delete a random beam                & Requires connectivity check                   & 15\% \\
    Split beam      & Insert node at beam midpoint        & Creates two beams, interpolates radii          & 15\% \\
    Randomize radii & Perturb all beam radii independently & Per-beam factor $\sim e^{\mathcal{N}(0, 0.15)}$ & 10\% \\
    \bottomrule
  \end{tabular}
\end{table}

\clearpage
\section{Out-of-distribution Detection and Confidence Head}
\label{app:ood_detection}

\paragraph{Candidate generation.}
We generate 7,981 mutated MicroPlate candidates from the 55 training topologies in two passes. Pass~1 (3,080 candidates): unguided random walks of depth $D=55$ from each training plate (55 seeds $\times$ 56 entries per walk = 3{,}080), using the mutation operators of Table~\ref{tab:mutations} with the same composition rules as the design search of Appendix~\ref{app:mutations}. Pass~2 (4,901 candidates): a 55-seed novelty-search variant with a shared global anchor list. Each training plate seeds an independent beam (width $B=2$); at each of $D=8$ iterations every beam member is expanded by $K=6$ aggressive mutations (perturb $\sigma=0.3$, scale $\sigma=0.5$, plus a $k$-node subgraph-replace operator with $k\in\{2,3,4\}$). Per-seed top-$B$ selection uses k-NN distance ($k=15$) to a global anchor list seeded with the 3{,}080 Pass~1 latent projections (the 55 training plates plus their 3{,}025 random-walk mutants) and updated with every Pass~2 evaluated candidate; this couples the per-seed walks so that seeds in dense regions migrate into unexplored directions. Of the 7,981 candidates, 6,919 obtained valid FEM characterization under the protocol of Appendix~\ref{app:mutations} (1,062 diverged at the maximum strain level); the latter retain features but have no $\rho$ label and contribute only to inference-time visualization.

\paragraph{Reconstruction error.}
For each candidate we encode the zero-displacement field (rest configuration), decode it back, and compute the mean squared displacement $\langle\|\hat{\mathbf{u}}_0\|^2\rangle$, where $\hat{\mathbf{u}}_0 = \mathrm{decode}(\mathrm{encode}(\mathbf{0}))$. This scalar measures how well the tokenizer round-trips the rest configuration on a given mesh.

\paragraph{Encoder-cycle inconsistency.}
For each candidate we encode the rest configuration to $\mathbf{z}_0$, then run $T=10$ autoregressive steps with action $[1,0,0,0]$ (uniaxial stretch). At each step $t$ the dynamics-predicted latent $\mathbf{z}_t$ is decoded to a displacement field $\mathbf{u}_t$, subsampled to the same $S=10{,}240$ FPS-coarsened query points used at encoding time, and re-encoded as a 3-frame window $(\mathbf{u}_{t-2}, \mathbf{u}_{t-1}, \mathbf{u}_t)$ to produce $\hat{\mathbf{z}}_t$ (the leading two frames are zero-padded for $t<3$). The per-frame metric is $c_t = \|\mathbf{z}_t - \hat{\mathbf{z}}_t\| / \|\mathbf{z}_t\|$ ($L^2$ norm of the flattened latent); the per-candidate scalar in Fig.~\ref{fig:ood_landscape}(C) is the mean over $T=10$ frames.

\paragraph{Confidence head architecture.}
The head input is a 1{,}050-dimensional concatenation of: tokenizer mean-pooled latent ($\mathrm{mean\_pool}(\mathbf{z}_0)$, 1,024 dimensions); tokenizer reconstruction error (1 dimension); per-frame consistency vector $(c_1, \ldots, c_{10})$ (10 dimensions); 15-dimensional graph statistics (node and beam counts, spatial spread, beam lengths, radii, connectivity, matching the features used in Fig.~\ref{fig:microplate_pca}). The head is a 3-layer MLP: $\mathrm{LayerNorm} \to \mathrm{Linear}(1050, 256) \to \mathrm{GELU} \to \mathrm{Dropout}(0.2) \to \mathrm{Linear}(256, 128) \to \mathrm{GELU} \to \mathrm{Dropout}(0.2) \to \mathrm{Linear}(128, 20)$. The 20 outputs are softmax-normalised over a uniform partition of $[-0.05, 1.0]$ in $\rho$. The predicted scalar $\hat{\rho}$ in Fig.~\ref{fig:ood_landscape}(D) is the expectation of bin centers under the softmax.

\paragraph{Training procedure.}
Optimization uses Adam (learning rate $10^{-3}$, weight decay $10^{-4}$), batch size 256, up to 200 epochs with early stopping (patience 30) on validation cross-entropy, on a single H100. We train three input configurations for ablation: free signals only (reconstruction error + 10-frame consistency, 11 input dimensions), latent only (mean-pooled $\mathbf{z}_0$, 1,024 input dimensions), and the full hybrid (1,050 input dimensions). The full configuration is used for Fig.~\ref{fig:ood_landscape}(D).

\clearpage
\section{Additional Results}
\label{app:additional}

\begin{figure}[h!]
  \centering
  \includegraphics[width=0.45\textwidth]{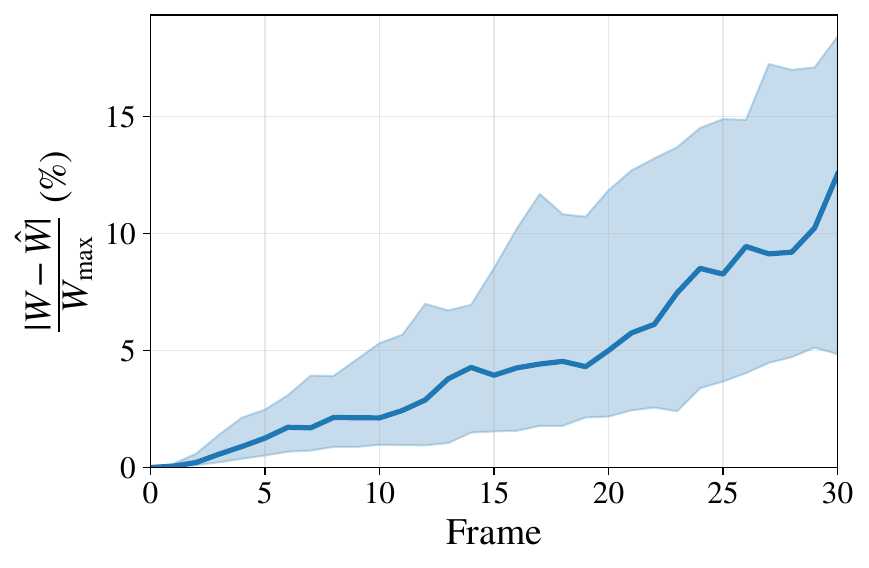}
  \caption{\footnotesize \textbf{Per-frame normalized work error on MicroPlate.} Per-frame normalized work error $|W - \hat{W}|/W_{\max}$ across 55 architected lattices during 30-step autoregressive rollout. Solid line: median. Shaded band: interquartile range.}
  \label{fig:rollout_curves}
\end{figure}

\begin{figure}[h!]
  \centering
  \includegraphics[width=0.4\textwidth]{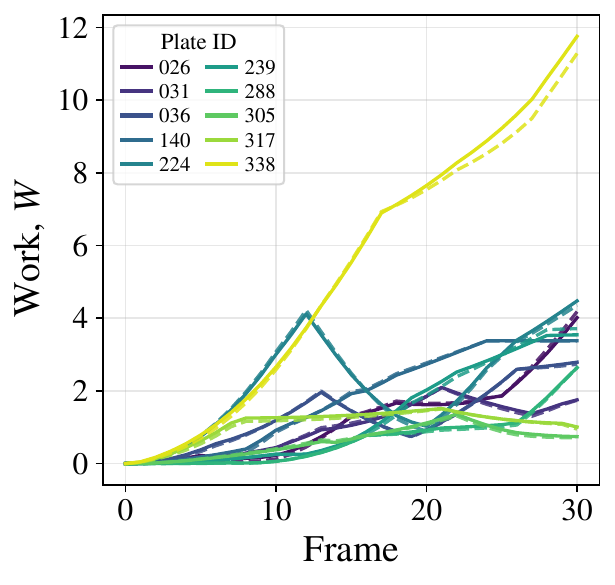}
  \caption{\footnotesize \textbf{Work vs.\ time on ten MicroPlate shapes.} Solid: FEM ground truth. Dashed: prediction.}
  \label{fig:fd_curves}
\end{figure}

\paragraph{Per-frame rollout curves.}
Figure~\ref{fig:rollout_curves} shows the per-frame normalized work error during 30-step autoregressive rollout across 55 architected lattices, and Figure~\ref{fig:fd_curves} shows example rollout work vs. time curves compared to ground truth.

\end{document}